\documentclass[conference]{IEEEtran}
\IEEEoverridecommandlockouts
\usepackage{cite}
\usepackage{amsmath,amssymb,amsfonts}
\usepackage{algorithmic}
\usepackage{graphicx}
\usepackage{textcomp}
\usepackage[hidelinks]{hyperref}
\usepackage{subcaption}
\usepackage{xcolor}
\usepackage{authblk}
\def\BibTeX{{\rm B\kern-.05em{\sc i\kern-.025em b}\kern-.08em
    T\kern-.1667em\lower.7ex\hbox{E}\kern-.125emX}}

\newcommand{\arrowtext}[1]{\mathrel{\stackrel{\text{#1}}{\longrightarrow}}}

\def\etal{\textit{et al.}}

\newcommand{\method}{AVACA}
\newcommand{\dataset}{MAVAD} 

\begin{document}
\bstctlcite{MyBSTcontrol}

\title{Audio-Visual Dataset and Method for Anomaly Detection in Traffic Videos}

\author[1]{Błażej Leporowski}
\author[1]{Arian Bakhtiarnia}
\author[2]{Nicole Bonnici}
\author[3]{Adrian Muscat}
\author[4,5]{\authorcr Luca Zanella}
\author[4]{Yiming Wang}
\author[1]{Alexandros Iosifidis}
\affil[1]{\textit{DIGIT, Department of Electrical and Computer Engineering, Aarhus University, Denmark} \authorcr {\tt \{bl, arianbakh, ai\}@ece.au.edu}\vspace{1.5ex}}
\affil[2]{\textit{Greenroads Ltd., TAKEOFF, University of Malta, Malta} \authorcr {\tt nicole@greenroadsmalta.com}\vspace{1.5ex}}
\affil[3]{\textit{Department of Computer Engineering, University of Malta, Malta} \authorcr {\tt adrian.muscat@um.edu.mt}\vspace{1.5ex}}
\affil[4]{\textit{Fondazione Bruno Kessler, Trento, Italy} \authorcr {\tt \{lzanella, ywang\}@fbk.eu}\vspace{1.5ex}}
\affil[5]{\textit{University of Trento, Trento, Italy}\vspace{1.5ex}}

\maketitle

\begin{abstract}
We introduce the first audio-visual dataset for traffic anomaly detection taken from real-world scenes, called \dataset, with a diverse range of weather and illumination conditions. In addition, we propose a novel method named \method \ that combines visual and audio features extracted from video sequences by means of cross-attention to detect anomalies. We demonstrate that the addition of audio improves the performance of \method \ by up to 5.2\%. We also evaluate the impact of image anonymization, showing only a minor decrease in performance averaging at 1.7\%.
\end{abstract}

\section{Introduction}
\label{sec:intro}

Detecting anomalies in videos has many real-world applications, such as surveillance. Video anomaly detection (VAD) can be succinctly described as the temporal and/or spatial localization of anomalous events in videos. VAD is an active area of research in deep learning literature. 
Most methods treat VAD as an unsupervised or weakly-supervised task where a significant amount of examples exist that represent normal scenes, and few outliers exist that represent anomalous data, typically only used for evaluation \cite{Mohammadi2021}. Similarly, the dataset and method in this work also treat video anomaly detection as a weakly-supervised binary classification problem where labels are only provided per video clip.

Methods that take advantage of multiple modalities have been shown to be successful in various deep learning tasks such as crowd counting \cite{hu2020} and action recognition \cite{Kazakos_2019_ICCV}. Despite the popularity of video anomaly detection as a research topic and the success of multimodal deep learning methods, few multimodal datasets exist for anomaly detection, and all available datasets are synthetically generated.
In contrast, we introduce an audio-visual anomaly detection dataset named Malta Audio-Visual Anomaly Detection (\dataset) with synchronized audio and video, collected from three locations in the Area. The dataset is anonymized by blurring faces and license plates to protect the privacy of individuals, and contains 11 classes, such as pedestrians, buses and heavy weight vehicles (trucks, vans, etc.).

Furthermore, we propose a novel audio-visual anomaly detection method called Audio-Visual Anomaly Cross-Attention (\method) that serves as a baseline for this dataset. Through extensive experiments, we demonstrate the improvements in performance resulting from the addition of audio features. Our dataset and code is publicly available\footnote{\url{https://gitlab.au.dk/maleci/audiovisualanomalydetection}}. 

\section{Related work}
\label{sec:related_work}

Over 30 publicly available datasets exist for video anomaly detection \cite{Kumari2021}, most notably, Shanghai Tech Campus Dataset \cite{liu2018ano_pred}, UCF-Crime \cite{sultani2018real} and UCSD-Pedestrian\footnote{\url{http://www.svcl.ucsd.edu/projects/anomaly/dataset.htm}} which are widely used in the literature. Video anomaly detection methods can be categorized into two classes: \textit{encoder-agnostic} methods and \textit{encoder-based} methods \cite{Feng2021}. Encoder-agnostic methods such as \cite{Wan2020} and \cite{Tian2021} use pre-trained visual feature extractors like I3D \cite{carreira2017quo}, X3D \cite{feichtenhofer2020x3d} and SlowFast \cite{feichtenhofer2019slowfast} and only train a classifier on extracted features. Encoder-based methods such as \cite{Zhong_2019_CVPR} train both the classifier and the feature extractor. From a different perspective, video anomaly detection datasets can be divided into three catogories: heterogeneous datasets that are diverse but do not contain specific anomalies; specific datasets that are bulit around anomalous events or objects; and other datasets that are not designed for anomaly detection but can still be useful for scene monitoring tasks \cite{Kumari2021}.

AVRL is an audio-visual method for detecting anomalies in crowds \cite{Gao2021}. AVRL utilizes pre-trained visual and audio feature extractors and feeds the extracted representations to an audio-visual fusion module, which is a simple MLP. AVRL is trained and tested using a synthetic dataset sourced from the video game Grand Theft Auto V. XD-Violence is an audio-visual dataset for violence detection, containing a large number 4,754 videos representing six classes of violent as well as peaceful behavior across various scenes \cite{wu2020not}. \cite{9354611} introduces an audio-visual anomaly detection method that extracts visual features, such as optical flow, and combines them with several extracted audio features, such as energy, volume and spectral flux, to determine anomalies. The authors also introduce a synthetic dataset created by adding gunshot sound effects taken from DCASE Challenge Task 2 \cite{mesaros2017dcase}, to the UMN dataset\footnote{\url{http://mha.cs.umn.edu/proj_events.shtml}} for unusual crowd activity detection in videos.

Three main types of multi-modal fusion approaches exist in the literature: early fusion, where the different modalities are combined via some operation and processed together from the beginning;
late fusion, where they are treated separately and are only combined in the very last layers; and intermediate fusion, where the modalities fuse within the network, and hence their features are learnt jointly \cite{Chumachenko2022}.

Self-attention, and by extension the transformer~\cite{NIPS2017_3f5ee243}, has quickly established itself as one of the core techniques in machine learning.
Not surprisingly, it has found wide application also in the problem of multi-modal fusion.
Transformer utilizes the concept of attention to capture the dependencies between different parts of the input sequence.
%
Within the context of fusing two modalities, $a_m$ and $b_m$, self-attention can be utilized by computing the query from modality $a_m$, and both keys and values from modality $b_m$.
This means that the final representation learned is of $a_m$ attending $b_m$, and further applying the calculated attention matrix to the features acquired from $b_m$.  
The output of the transformer, after passing the final representation though a fully connected layer, will be further on referred to as $\tau$.

\section{Dataset}
\label{sec:dataset}

\begin{figure*}
  \centering
  \includegraphics[width=\textwidth]{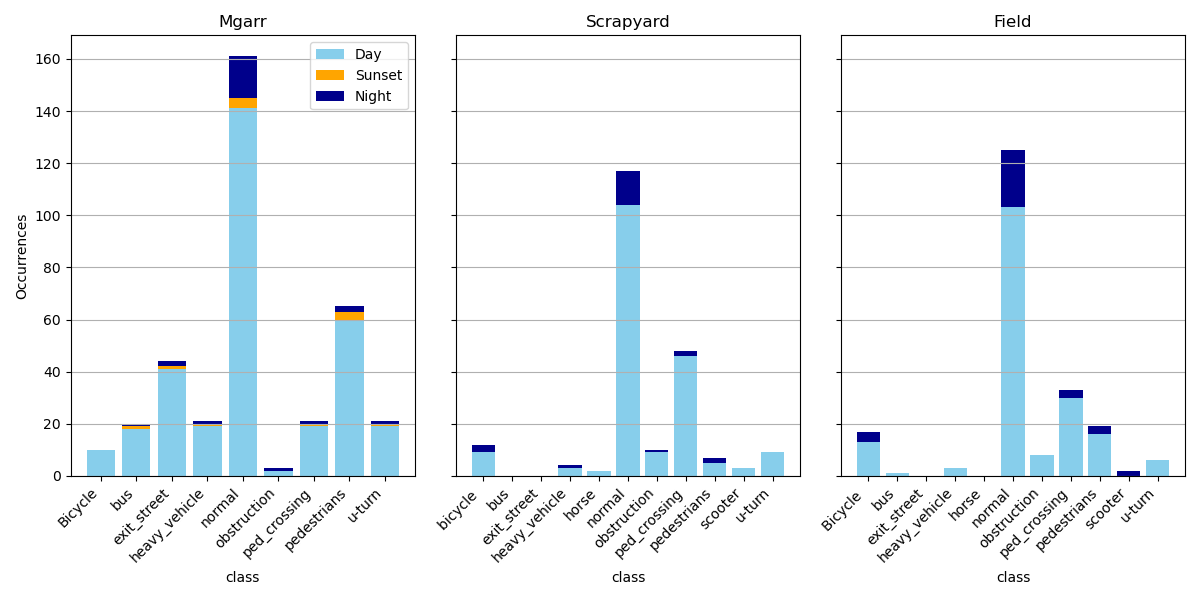}
  \caption{Weather and illumination conditions for the recorded videos.}
  \label{fig:weather_stats}
\end{figure*}

The proposed \dataset \ dataset consists of 764 videos gathered from three different locations.
Those locations have been chosen with the focus on road traffic, including various types of vehicles, bicycles and pedestrians. 
The raw audio and video data was collected from two locations on the island of Malta: one in Zejtun, a town close to the industrial region on the eastern coast, and another in Mgarr, a rural town on the western coast. 
Three audio-visual cameras were deployed, two in Zeytun, and one in Mgarr. 
The data is organized into 11 classes: normal situations, pedestrians, pedestrians crossing the street, vehicles exiting the side street, heavy weight vehicles (trucks, vans, etc.), buses, bicycles, obstructions, u-turns, scooters and horses. 
Videos that do not belong to any specific anomalous class are labeled as ``normal''. 
Table \ref{tab:dataset_stats} lists the number of videos pertaining to each of these classes per scene, while Fig.~\ref{fig:weather_stats} illustrates the weather and illumination conditions during the recordings.
Table \ref{tab:camera_params} presents the details of the hardware used for data collection. The specific camera models used were: Safire 5MP Bullet outdoor/indoor IP camera With POE for Mgarr, and ANNKE outdoor 5MP PoE security cameras, model I51DL, with 2.8mm lens for both Zeytun scenes. 
The data has been masked to prevent the recording of private property. Firstly through the use of camera configuration, secondly with a hand crafted mask overlayed on the images, so that only the public road is visible. 


\begin{table}[h]
\centering
\caption{Dataset statistics.}
\label{tab:dataset_stats}
\resizebox{0.48\textwidth}{!}{
\begin{tabular}{|c|c|c|c|}
\hline
Label & Mgarr & Zejtun Scrapyard & Zejtun Field \\
\hline
normal & 161 & 117 & 125 \\
pedestrians & 65 & 7 & 19 \\
pedestrians crossing the road & 21 & 48 & 33 \\
bicycle & 10 & 12 & 17 \\
bus & 20 & 0 & 1 \\
exit side street & 44 & 0 & 0 \\
heavy goods vehicle & 21 & 4 & 3 \\
obstruction & 3 & 10 & 8 \\
u-turn & 1 & 9 & 6 \\
scooter & 0 & 3 & 2 \\
horse & 0 & 2 & 0 \\
\hline
Total & 346 & 212 & 214 \\
\hline
\end{tabular}
}
\end{table}

\begin{table*}[h]
\centering
\caption{Camera parameters}
\label{tab:camera_params}
\resizebox{0.7\textwidth}{!}{
\begin{tabular}{|c|c|c|c|}
\hline
Specifications type & Mgarr & Zejtun Scrapyard & Zejtun Field \\
\hline
Resolution & 1920x1080 (2.07 MP) & 1280x720 (0.92 MP) & 1920x1080 (2.07 MP) \\
Non-Masked Resolution & 0.61 MP & 0.28 MP & 0.56 MP \\
Framerate & 25 & 30 & 30 \\
Audio encoding & PCM 16-bit & PCM 16-bit & PCM 16-bit \\
Audio sampling rate & 48kHz & 48kHz & 48kHz \\
Audio channels & 2 & 2 & 2 \\
\hline
\end{tabular}
}
\end{table*}

Following the taxonomy in~\cite{Kumari2021}, this dataset fulfills the requirements for both heterogenous and specific, since it contains varied and well documented sets of anomalies provided in a weakly-labeled manner. That is, each anomalous clip represents almost exclusively the anomaly, and the dataset provides per clip labels. Therefore, the dataset can be used both for unsupervised and weakly-supervised methodologies, and provides the opportunity for binary or multi-class detection.

\begin{figure*}
\setlength{\fboxrule}{0pt}
\begin{center}
\begin{tabular}{ l c c }
Scene & Video Frame & Spectrogram\\
\hline
Mgarr &
\includegraphics[height=4cm]{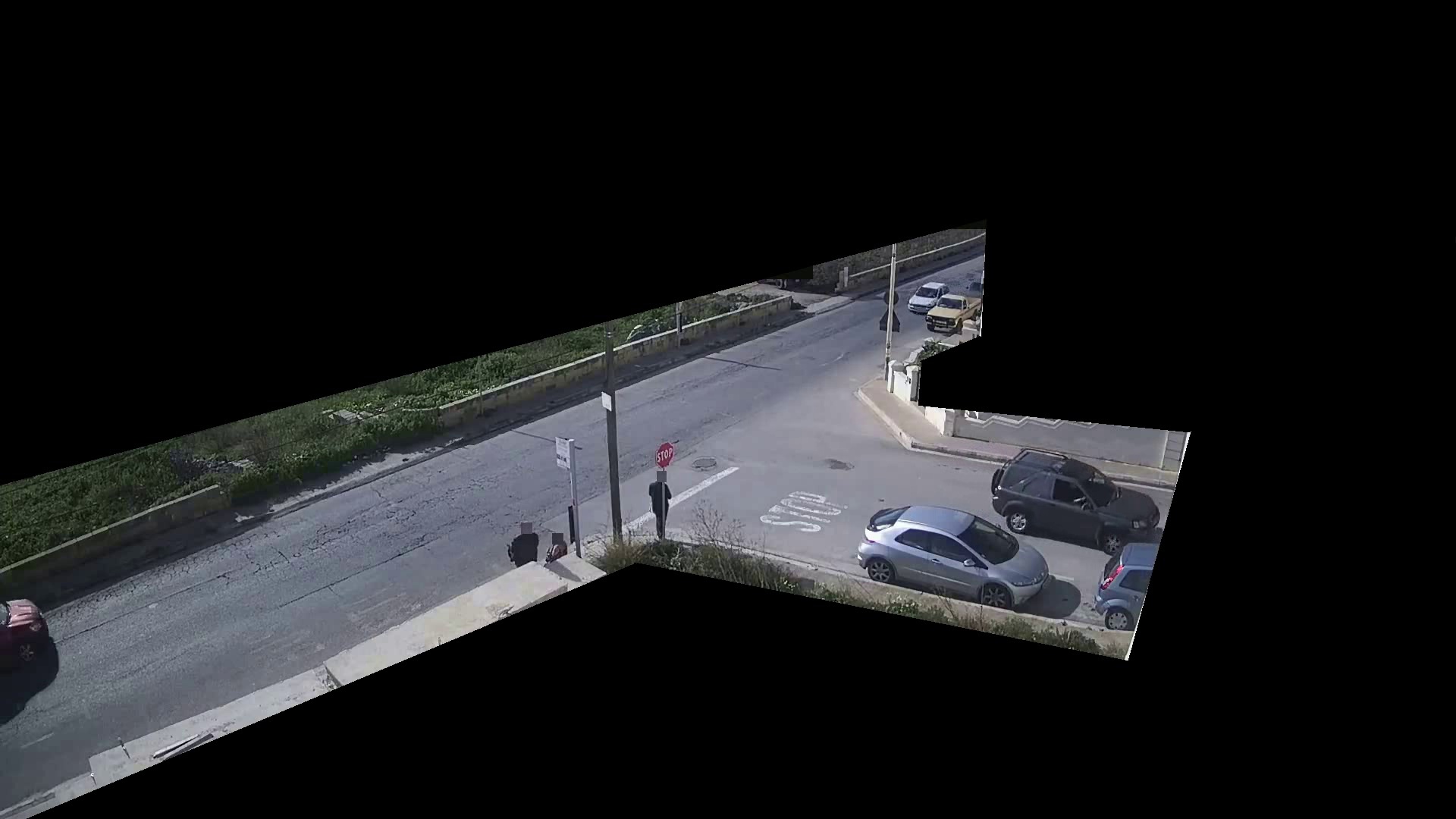} &
\includegraphics[height=4cm]{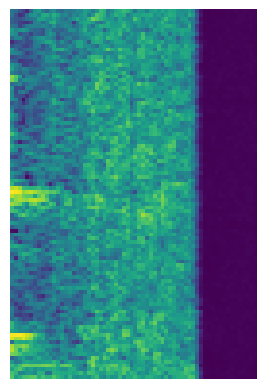}\\
Zejtun Scrapyard &
\includegraphics[height=4cm]{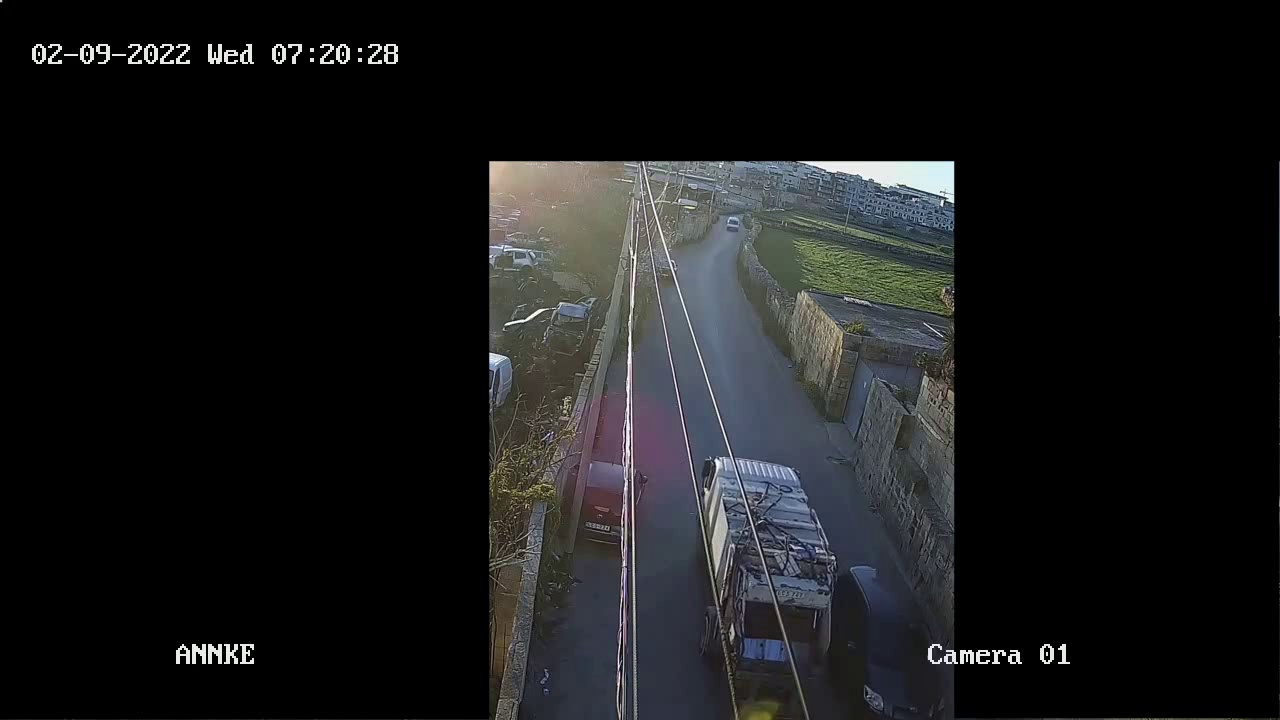} &
\includegraphics[height=4cm]{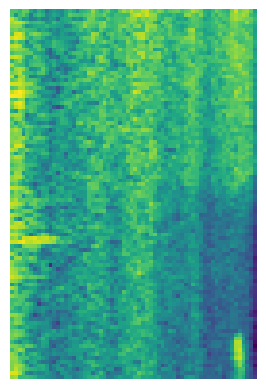}\\
Zejtun Field &
\includegraphics[height=4cm]{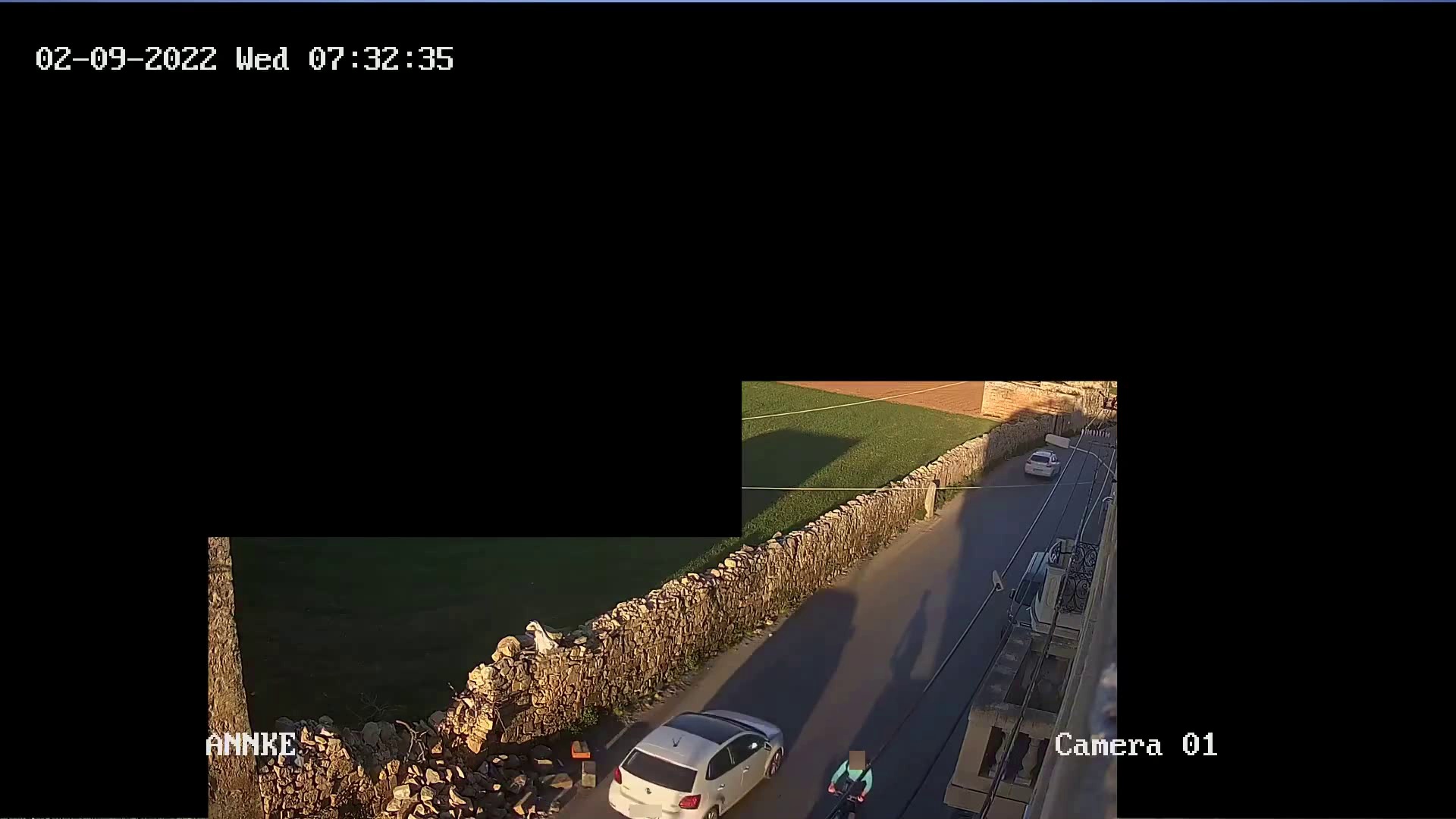} &
\includegraphics[height=4cm]{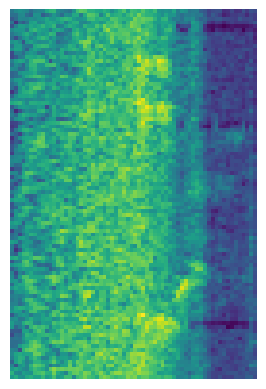}\\
\end{tabular}
\end{center}
\caption{Example video frames and audio spectrograms from Mgarr, Zejtun Scrapyard and Zejtun Field scenes.}
\label{fig:scene_examples}
\end{figure*}

\subsection{Data anonymization}

\begin{figure}[htbp]
    \centering
    \includegraphics[width=0.48\textwidth]{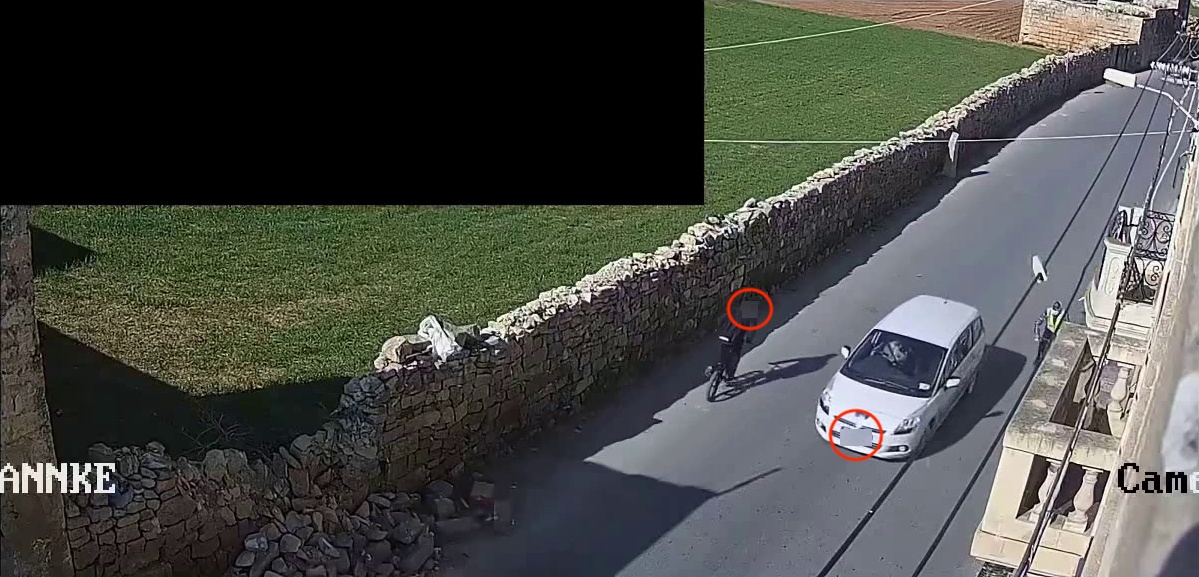}
    \caption{Example anonymized frame from the Zejtun Field
 scene. Blurred faces and license plates are highlighted by red circles.}
    \label{fig:field_example}
\end{figure}

The proposed dataset captures public scenes, therefore, it is not feasible to obtain consent forms from everyone appearing in the videos.  To protect the privacy of the general public and be compliant with European regulations including GDPR, we anonymize the raw videos to remove personal identifiable information, including faces of persons and license plates of vehicles, as shown in Figure \ref{fig:field_example}. Anonymization is achieved by first detecting the faces and license plates on each video frame and then applying Gaussian blur to the detected regions. We use the general-purpose object detector YOLOv5\footnote{\url{https://github.com/ultralytics/yolov5}}, pre-trained on the MS COCO dataset \cite{lin2014microsoft}, to detect faces and license plates. However, when applied out-of-the-box to our surveillance videos, it fails to detect most of the faces and license plates due to strong domain shifts and frequent occlusions. 
To mitigate this issue, we fine-tuned the face detector using CrowdHuman~\cite{shao2018crowdhuman}, a publicly available dataset containing CCTV surveillance footage. We also fine-tuned the license plate detector with annotated videos captured from Area to achieve the best detection accuracy. The audio data is maintained as it is, without anonymization, because the microphone is positioned at a distance and is unlikely to capture any personal identifiable information regarding voice and speech.

\section{Audio-Visual Anomaly Detection Method}
\label{sec:method}
This section begins with the problem statement, and follows up with the general description of the proposed architecture.
Next, the used fusion technique, the employed losses, and their impact on the model's behaviour are explained in more detail.

\subsection{Problem statement} 
The proposed \method \ training regimen is based on the work of Wan \etal~\cite{Wan2020}. 
As such, to fully understand the impact that the Dynamic Multiple-Instance Learning Loss and the regularizing Center loss have on the model's training behaviour, the notation introduced in the problem statement follows that proposed by Wan \etal. 
A training set consisting of $n$ videos is denoted by $\mathcal{X} = \{x_i\}_{i=1}^n$. Each video is temporally split in a $t_i$ video clips, for instance, based on a rolling window size and stride, which may depend on the chosen feature extractor. 
The set of anomaly labels of the training videos is denoted as $Y = \{y_i\}_{i=1}^n$, where $y_i = \{0, 1\}$. 
In the test phase, the predicted anomaly score vector of a video $x$ is denoted as $s = \{s^{j}\}_{j=1}^t$, where $s^{j} \in [0, 1]$ is the anomaly score of the $j$-th video clip.

\subsection{Architecture overview}
\label{subsec:arch_overview}
From a high level perspective, the proposed \method \ method consists of two processing paths: audio and visual.
Each path starts with its respective, pretrained feature extractor.
Each path has two learning stages, and a transformer layer placed in between those stages.
The transformer placed in the visual path is further called the VAT, and the transfomer placed in the audio path is further called AVT.
On a conceptual level, the input to the model is an audio-visual sequence, and its output is the predicted anomaly score vector. 
The architecture can be seen in Fig.~\ref{fig:avad_architecture}, details of which are further explained in the following sections.

\subsection{Feature extractors} 
The proposed method belongs to the category of encoder-agnostic methods, that is, the pretrained feature extractors are employed without any further training nor finetuning.
Feature extractors are tasked with acting as the first step in the model.
They take as input raw data, video frames and audio snippets in our case, and transform them into a new, distilled representation.
Feature extractors are often methods that have been trained on large scale, industry leading datasets such as the Kinetics~\cite{kay2017kinetics} dataset or the AudioSet~\cite{gemmeke2017audio} dataset.
In the context of feature extractors, the part of the model that learns generalized representations is refered to as a 'body', and the specialized part, for example classifier, as the 'head'.
When trained on large scale datasets, the bodies of those methods are well suited to serve as general purpose feature extractors.

The choice of the feature extractors determines also the exact length of the input sequence.
We have narrowed our choice of visual feature extractors between SlowFast~\cite{feichtenhofer2019slowfast}, which takes as input 32 frames, and X3D~\cite{feichtenhofer2020x3d}, which has been tested on 16-frame long inputs. 
Both models have been pretrained on the Kinetics dataset.
The audio extractor chosen is the widely adopted VGGish~\cite{hershey2017cnn}, which has been trained on the AudioSet.
Our initial experiments showed that SlowFast achieved better results than X3D.
We have thus chosen SlowFast, specifically SlowFast\_r50, as our visual feature extractor.

The selection of SlowFast imposes the requirements upon the input length and resolution.
SlowFast requires two inputs: the slow and fast paths.
The fast path is a sequence of 32 images denoted as \( F = \{ v_i \}_{i=1}^{32} \), where each \( v_i \) is a 3-channel image.
The slow path, denoted as \( S = \{ s_i \}_{i=1}^{8} \), is a sequence of 8 images created by selecting every fourth frame from the original 32-frame sequence.
The resolution of the frame's short side is set to 256, while the long edge is kept in scale. 
The sequence is then normalized. 

The input to the audio feature extractor, VGGish, is a single spectrogram \( a_i \).
However, as Slowfast imposes the requirement to process 32 frames for a single sequence, the resulting input to the VGGish model is a sequence of 32 spectrograms denoted as \( A = \{ a_i \}_{i=1}^{32} \). 
During inference, VGGish processes the spectrograms consecutively.
These spectrograms are created from 32 overlapping audio snippets, each 1 second long. 
Each audio snippet starts 1 second before and ends exactly at the time of its respective frame. 
The audio snippets are generated using a rolling window applied to the audio file, with a stride value set to ensure the synchronization with the frame rate of the videos. 
The spectrograms are created in the standard way as required for the VGGish network, specifically using Mel spectrogram \cite{choi2016automatic}.

The resulting multimodal input to the model is a tuple denoted as \( M = ((F, S), A) \).
The output of the visual and audio feature transformers are the feature matrices \( {V}_i \) and \( {P}_i \), respectively,  composed of features from the training video \( x_i \).

\subsection{Architecture details}

The architecture and implementation is based on the multimodal data fusion approaches proposed in~\cite{Chumachenko2022}.
As previously stated in Section~\ref{subsec:arch_overview}, the fusion module consists of two paths, one for visual and one for audio processing.
Each path contains 2 stages, each in turn containing 2 layers.
The first stages of the paths serve the purpose of finetuning the transformed input features during the model training.
Stage 1 contains 2D convolutions for the visual path, and 1D convolutions for the audio path.
The first stage transformation can be presented as \( ({V}_i, {P}_i) \arrowtext{Stage 1} ({V}_i', {P}_i')\).
 
Next, we include two transformers, one for each path.
The transformers, placed in between the stages, are identical and are meant to enhance the separate paths with the knowledge distilled from the other modality.
However, with 32 overlapping audio segments, a substantial amount of the information carried by the audio track is repeated, and crucial, yet more subtle changes in the recordings may not get picked up by the model.
Hence, we introduce a simple operation that creates two different audio inputs for the two transformers.
The plain audio sequence can be described as $\mathbf{{P}_i'} = ({p}_1', {p}_2', \ldots, {p}_{32}')$, with ${p}_i'$ representing the $i$th item.
The resulting sequence, obtained by subtracting the previous item from each element, is denoted as $\mathbf{U} = (u_1, u_2, \ldots, u_{31})$.
The operation can be described as $u_i = {p'}_{i+1} - {p'}_i$ for $i = 1, 2, \ldots, 31$.

The transformer in the visual path, VAT, aims to enhance the visual representation of the video with the full audio information.
It has thus the following input: $K_{VAT} = {V'}_i$, $V_{VAT} = {V'}_i$, $Q_{VAT} = {P'}_i$.
The second transformer, AVT, is placed in the audio path.
The VA transformer is queried by the video patches, and as keys and values takes the modified audio patches, which makes: $K_{AVT} = {U}_i$, $V_{AVT} = {U}_i$, $Q_{VAT} = {V'}_i$. 
This procedure aims to differentiate the two paths further, that is, the VAT looks at how the full audio enhances the visual signal, and the AVT focuses on the changes in the audio path and their relationship with the visuals.
When $K_{VAT} = {U}_i$ and $V_{VAT} = {U}_i$, the procedure will be further called focused audio path.
When $K_{VAT} = {P'}_i$ and $V_{VAT} = {P'}_i$, the procedure will be further called plain audio path.

\begin{figure*}[htbp]
    \centering
    \includegraphics[width=\textwidth]{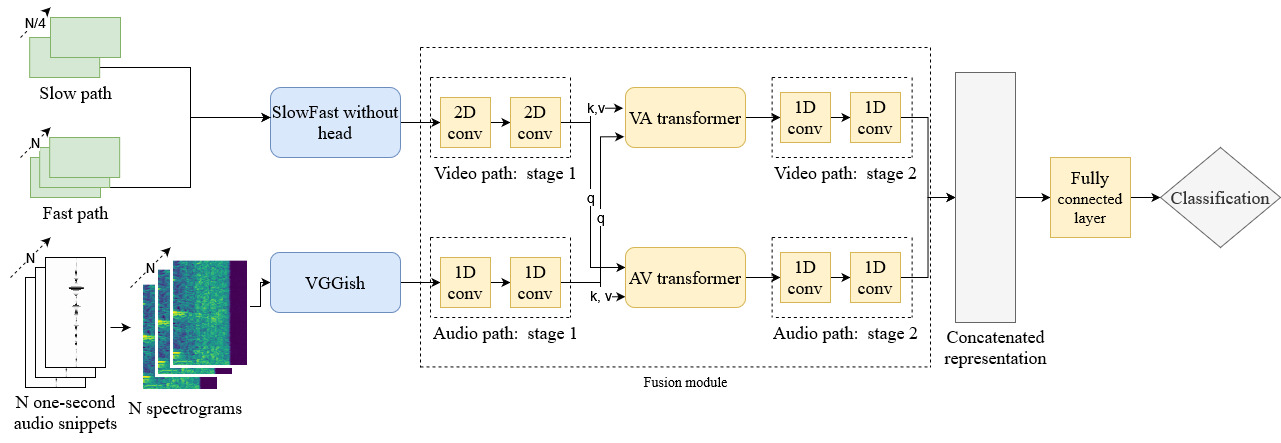}
    \caption{The \method \ architecture.}
    \label{fig:avad_architecture}
\end{figure*}

Finally, stage 2 consists of 1D convolutions for both branches.
The second stage transformation can be presented as \( ({\tau}_{VAT}, {\tau}_{AVT}) \arrowtext{Stage 2} ({\tau}_{VAT}', {\tau}_{AVT}') \).
The outputs of the second stages are concatenated, and passed to a fully connected layer that outputs a binary vector $s_j$. 
The complete architecture can be seen in Fig.~\ref{fig:avad_architecture}.

\subsection{Training losses}
During training, the model aims to minimize 2 losses, taken from Wan \etal~\cite{Wan2020}: the dynamic multiple-instance learning loss ($L_{DMIL}$), and the center loss ($L_C$).
$L_{DMIL}$ is designed to enlarge the inter-class distance between anomalous and normal instances, while $L_C$ `pulls' in the opposite direction, minimizing the intra-class distance of normal instances. 

\textbf{Dynamic Multiple-Instance Learning (DMIL): }In the context of Multiple Instance Learning, a positive bag contains at least one positive instance, while a negative bag contains no positive instances. 
For anomaly detection in videos, abnormal videos have at least one anomalous event while normal videos have no anomalous events. 
To enhance the distinction between anomalous and normal instances with weak supervision, Wan \etal introduced the DMIL loss, taking into consideration the diversity in video duration. 
They also introduced the k-max selection method to obtain the k-max anomaly scores.
The value of k is determined based on the number of clips in a video, given by $k_i = \left\lfloor \frac{t_i}{\alpha} \right\rfloor$, where $\alpha$ is a hyperparameter. Thus, the k-max anomaly scores of the $i$-th video can be represented as $S_i = \{ p_{i}^j \,|\, j = 1, 2, \ldots, k_i \}$, where $p_i = \text{sort}(s_i)$,
$s_i$ is the anomaly score vector of the $i$-th video, $\text{sort}(\cdot)$ is a descending sort operator, and $S_i$ consists of the top-$k_i$ elements in $s_i$. The DMIL loss can then be represented as:
\begin{equation}
    L_{\text{DMIL}} = \frac{1}{k_i} \sum_{s_i^j \in S_i} \left[ -y_i \log(s_{i}^j) + (1 - y_i) \log(1 - s_{i}^j) \right], \quad
\end{equation}
where $y_i = \{0, 1\}$ is the video anomaly label. 
Next, the authors calculate the cross-entropy between each of the selected $k$ scores and the video label as the instance loss, respectively. 
Noise labels can affect the anomaly scores of the sample features from which an average anomaly score is calculated.
However, the DMIL loss focuses on individual anomaly scores rather than an average one, thereby preventing propagation of errors brought by noise labels.

\textbf{Center loss: } The objective of the DMIL loss is to enlarge the inter-class distance of instances. 
However, both the max and k-max selection methods inevitably produce wrong label assignments, especially in the early training stages when the anomaly scores of normal clips and abnormal clips in an abnormal video are similar. 
This leads to the enlargement of the intra-class distance of normal instances by the DMIL loss, which can reduce detection accuracy in the testing stage. 
Wan \etal propose a center loss for anomaly score regression to address this issue. 
This loss focuses on gathering the anomaly scores of normal video clips:
\begin{equation}
    L_c =
    \begin{cases}
    \frac{1}{t_i}\sum_{j=1}^{t_i}{||s_{i}^j - c_i||}^2_2, & \text{if } y_i = 0, \\
    0, & \text{otherwise,}
    \end{cases}
    \quad
\end{equation}
where $c_i$ is the center of the anomaly score vector $s_i$ of the $i$-th video, that is, $c_i = \frac{1}{t_i}\sum_{j=1}^{t_i} s_{j}^j$. 
The total loss function is thus $L = \theta L_{DMIL} + \lambda L_c$, where $\theta$ and $\lambda$ are hyper-parameters.

\section{Experiments}
\label{sec:experiments}

In all experiments we prepare the video and audio features first, to facilitate faster training times.
In this case, the feature encoders are unused during training, validation and testing.
Once the model is deployed and starts operating in a real-world scenario, they are activated.

For each scene in the \dataset \ dataset, we repeat the experiment 3 times with a different randomly split data. We use $60\%$ of the data for training, $20\%$ as validation and $20\%$ as test data.
The annotations generator aims to apply the $60/20/20$ split to each class, if possible.
This results in balanced inter-class distribution, yet randomized intra-class assignments.

\subsection{Baseline}
\label{subsec:experiments_baseline}
In order to allow for relating the performance of the proposed method and the results achieved on the proposed dataset, we conducted an experiment on the Shanghai Tech dataset.
As this dataset does not contain audio, we create null audio signals in the form of tensors filled with zeroes.
This means that the audio branch carries no information, and the model deals purely with the visual features.
However, this allows us to use the same architecture for all tests and create a reliable comparison.
We set the learning rate to $10^{-5}$, and use $\lambda = 1$ and $\theta = 20$. 
Table \ref{tab:Shanghai Tech_results} presents the results achieved by \method \ on the Shanghai Tech dataset in comparison to the current SotA. \method \ achieves a performance that is competitive with currently best performing models on the Shanghai Tech dataset. 
To extend this baseline to the proposed \dataset \ dataset, we follow the same procedure for all of them.


\subsection{Results}

Table \ref{tab:avad_results} presents the complete results of applying \method \ on the \dataset dataset.
The loss weights are the same for all cases: $\lambda = 1$, and $\theta = 10$. 
All models have been trained for 100 epochs with a starting learning rate of $10^{-5}$ and 4 attention heads in each transformer.

\begin{table}[h]
\centering
\caption{Results on \dataset\ dataset (AUC).}
\label{tab:avad_results}
\resizebox{0.48\textwidth}{!}{
\begin{tabular}{|l|c|c|c|}
\hline
Scene & Zeroed audio & Focused audio path & Plain audio path \\
\hline
Mgarr & $87.91 \pm 3.47\%$ & $89.76 \pm 1.33\%$ & $88 \pm 1.82\%$ \\
Zejtun Scrapyard & $63.18 \pm 5.85\%$ & $64.10 \pm 6.07\%$ & $64.69 \pm 4.4\%$ \\
Zejtun Field & $76.58 \pm 13.09\%$ & $80.45 \pm 9.73\%$ & $78.6 \pm 6.82\%$ \\
\hline
\end{tabular}
}
\end{table}

\begin{table}[h]
\centering
\caption{Results on Shanghai Tech}
\label{tab:Shanghai Tech_results}
\begin{tabular}{|c|c|}
\hline
Method & AUC ($\%$) \\
\hline
AR-Net~\cite{Wan2020} & 91.24\\
\textbf{\method \ (ours)} & \textbf{93.79} \\
MIST~\cite{Feng2021} & 94.83 \\
RTFM~\cite{Tian2021} & 97.21 \\
S3r~\cite{Li2022} & 97.48 \\
SSRL~\cite{Wu} & 97.98 \\
\hline
\end{tabular}
\end{table}

\begin{table}[h]
\centering
\caption{\dataset\ raw vs anonymized}
\label{tab:avad_anonymization_impact}
\begin{tabular}{|l|c|c|}
\hline
Scene & Raw data & Anonymized data \\
\hline
Mgarr & $91.49 \pm 1.29\%$ & $89.76 \pm 1.33\%$ \\
Zejtun Scrapyard & $66.5 \pm 4.61\%$ & $64.10 \pm 6.07\%$ \\
Zejtun Field & $80.29 \pm 9.01\%$ & $80.45 \pm 9.73\%$ \\
\hline
\end{tabular}
\end{table}

For all 3 scenes, the addition of audio improves performance.
For the Mgarr scene, the absolute AUC improvement is $1.85\%$, for the Zejtun Scrapyard scene $0.85\%$ and for the Zejtun Field scene $3.87\%$.
By looking at Fig.~\ref{fig:scene_examples}, we can see that the Zejtun Scrapyard scene  differs from the other scenes.
The street that this camera observers runs vertically in the camera view, meaning that the camera captures changes far into its field of view, while the microphone can be too far to record the relevant audio, or simply the noise from occurrences closer to the camera may obscure the audio relevant to the far off view.
This explains why addition of audio in the Zejtun Scrapyard scene does not improve the performance in a significant way.
The Zejtun Scrapyard scene is also the only scene where using the focused audio path results in worse performance.
This is, again, probably down to the fact that the information from video and camera is less correlated.


\subsection{Impact of Anonymization}
We also performed an anaylsis of the impact that image anonymization has on the performance of the proposed method.
Due to privacy concerns only the anonymized version of the dataset can be made public, but in Table~\ref{tab:avad_anonymization_impact} we present the comparison of results achieved by the \method \ using the focused path audio applied to the exact same splits of data on the raw and anonymized versions of data.


Across the 3 scenes, the relative change in performance after anonymization is $-1.7\%$.
The biggest drop in performance is in the case of the Zejtun Scrapyard scene, while the Zejtun Field scene even notes a small improvement. 
The anonymization process leads to a small reduction of the sharpness of the images but, in general, the impact can be described as not significant.

\section{Conclusion}
\label{sec:conclusion}

A novel audio-visual anomaly detection dataset with 3 scenes was proposed, filling a crucial gap in the pool of publicly available data.
Furthermore, the proposed audio-visual anomaly detection method, \method, showcases that the addition of audio can indeed improve performance and reduce deviation in the results.
The proposed method was also benchmarked against the popular, but visual only, Shanghai Tech dataset by creating null (zeroed) audio signal and is competitive.

\section*{Acknowledgement}

This work was funded by the European Union’s Horizon 2020 research and innovation programme under grant agreement No 957337.

\bibliographystyle{IEEEtran}
\bibliography{references.bib}

\end{document}